\documentclass[a4paper,11pt]{article}

\usepackage{graphicx}  
\usepackage{url}       
\usepackage{times}
\usepackage{times}
\usepackage{latexsym}
\usepackage{csquotes}
\usepackage{graphicx}
\usepackage{subcaption}
\usepackage{footmisc}

\usepackage{url}
\usepackage{natbib}
\usepackage[margin=25mm]{geometry}

\title{The Pragmatics of Indirect Commands in Collaborative Discourse}
\date{}

\author{Matthew Lamm\thanks{\hspace{0.5em} Authors contributed equally} \\
  Stanford Linguistics \\
  Stanford NLP Group\\
  {\tt mlamm@stanford.edu} \and
  Mihail Eric\footnotemark[1] \\
  Stanford Computer Science \\
  Stanford NLP Group\\
  {\tt meric@cs.stanford.edu} \\}

\begin{document}
\maketitle
\thispagestyle{empty}
\pagestyle{empty}

\begin{abstract}
Today's artificial assistants are typically prompted to perform tasks through direct, imperative commands such as \emph{Set a timer} or \emph{Pick up the box}. However, to progress toward more natural exchanges between humans and these assistants, it is important to understand the way non-imperative utterances can indirectly elicit action of an addressee. In this paper, we investigate command types in the setting of a grounded, collaborative game. We focus on a less understood family of utterances for eliciting agent action, locatives like \emph{The chair is in the other room}, and demonstrate how these utterances indirectly command in specific game state contexts. Our work shows that models with domain-specific grounding can effectively realize the pragmatic reasoning that is necessary for more robust natural language interaction.
\end{abstract}

\section{Introduction}
A major goal of computational linguistics research is to enable organic, language-mediated interaction between humans and artificial agents. In a common scenario of such interaction, a human issues a command in the imperative mood---e.g. \emph{Put that there} or \emph{Pick up the box}---and a robot acts in turn \citep{bolt1980put,tellex2011understanding,walter2015situationally}. While this utterance-action paradigm presents its own set of challenges \citep{tellex2012toward}, it greatly simplifies the diversity of ways in which natural language can be used to elicit action of an agent, be it human or artificial \citep{clarkusinglanguage1996,portner2007imperatives,kaufmann2009unified,condoravdi2012imperatives,kaufmann2016fine}. Most clause types, even vanilla declaratives, instantiate as performative requests in certain contexts \citep{austin1975things,searle1989performatives,perrault1980plan}.

In this work, we employ machine learning to study the use of performative commands in the Cards corpus, a set of transcripts from a web-based game that is designed to elicit a high degree of linguistic and strategic collaboration \citep{Djalali-etal:2011,Djalali:Lauer:Potts:2012,Potts:2012WCCFL}. For example, players are tasked with navigating a maze-like gameboard in search of six cards of the same suit, but since  a player can hold at most three cards at a time, they must coordinate their efforts to win the game.

We focus on a subclass of performative commands that are ubiquitous in the Cards corpus: Non-agentive declaratives about the locations of objects, e.g. ``The five of hearts is in the top left corner," hereafter referred to as \textit{locatives}. Despite that their semantics makes no reference to either an agent or an action---thus distinguishing them from conventional imperatives \citep{condoravdi2012imperatives}---locatives can be interpreted as commands when embedded in particular discourse contexts. In the Cards game, it is frequently the case that an addressee will respond to such an utterance by fetching the card mentioned.

Following work on the context-driven interpretation of declaratives as questions \citep{beun2000context}, we hypothesize that the illocutionary effect of a locative utterance is a function of contextual features that variably constrain the actions of discourse participants. To test this idea, we identify a set of 94 locative utterances in the Cards transcripts that we deem to be truly ambiguous, out of context, between informative and command readings. We then annotate their respective transcripts for a simplified representation of the tabular common ground model of \cite{malamud2015three}. Here, we identify the common ground with the state of a game as reflected by the utterances made by both players up to a specific point in time. Finally, we train machine learning classifiers on features of the common ground to predict whether or not the addressee will act in response to the utterance in question. Through these experiments we discover a few very powerful contextual features that predict when a locative utterance will be interpreted as a command.

\section{Related Work}

The subject of indirect commands, of which the locative utterances we study are an example, has been extensively analyzed in terms of speech act and decision theory \citep{austin1975things,clark1979responding,perrault1980plan,allen1980analyzing,searle1989performatives}. In \citeauthor{portner2007imperatives}'s (2007) formal model, imperatives are utterances whose conventional effect updates an abstract ``to-do list" of an addressee. More recent debate has asked whether this effect is in fact built into the semantics of imperatives, or if their directive force is resolved by pragmatic reasoning in context \citep{condoravdi2012imperatives,kaufmann2016fine}. 

The present work synthesizes these intuitions from the theory of commands with recent computational work on natural language pragmatics \citep{Vogel:Potts:Jurafsky:2013,Vogel-etal:2014,degen2013cost} and collaborative dialogue \citep{chai2014collaborative}. We are particularly influenced by previous work demonstrating the complexity of pragmatic phenomena in the Cards corpus \citep{Djalali-etal:2011,Djalali:Lauer:Potts:2012,Potts:2012WCCFL}.

\section{The Cards corpus}

The Cards corpus is a set of 1,266 transcripts from a two-player, collaborative, web-based game. The Cards corpus is well-suited to studying the pragmatics of commands because it records both utterances made as well as the actions taken during the course of a game.

At the start of the game, player locations are randomly initialized on a two-dimensional, grid-style game board. Cards from a conventional 52-card deck are scattered randomly throughout the board. Players are prompted with the following task:

\begin{displayquote}
Gather six consecutive cards of a particular suit (decide which suit together), or determine that this is impossible. Each of you can hold only three cards at a time, so you’ll have to coordinate your efforts. You can talk all you want, but you can make only a limited number of moves.
\end{displayquote}

In addition to the fact that players can only hold three cards at a time, the game is further-constrained in ways that stimulate highly collaborative talk. In particular, while players can see their own location, they cannot see the locations of their partners and so must inquire about them. Players can only see cards within a small neighborhood around their respective locations, and so must explore the board to find relevant cards. Moreover, while some walls are visible, others are invisible and so lead to surprise perturbations in the course of exploring the gameboard. 

\section{Command types in the Cards Corpus}

Commands in the Cards corpus can be coarsely divided into ones which make reference to an action with a second person agent, and those which do not. 

The first of these categories is comprised of imperatives and a variety so-called performative commands: Utterances  which act as commands in context but whose clause type is not conventionally associated with the effect of commanding  \citep{clark1979responding,searle1989performatives,wierzbickacrossculturalpragmatics}. For example, with respect to picking up cards:
\begin{displayquote}
pick it up! \\
pick up the 9 \\
or hell, grab the 234 of D \\
ok when you get here pick up the 8H \\
i think you should pick up the 3h, 5h, and 8h \\
so if you can find 5S,6S,7S that would be great 
\end{displayquote}

\noindent With respect to dropping (or not dropping) cards:
\begin{displayquote}
drop the 2 \\
keepp the 3 \\
no dont drop it.) \\
ok drop the 7 i guess\\
get rid of 6d. i found 7h \\
so if you come and drop the 8h and pick up the 6h we are good 
\end{displayquote}

\noindent With respect to conversational actions (some of these utterances are shortened for clarity):
\begin{displayquote}
tell me where it is\\
talk to me dude [...]\\
tell me if you see 5 or 6 \\
don't just say ``a lot of cards" [...] \\
awesome let me know once you have it.

\end{displayquote}

Imperatives and performatives that mention agents contrast with the lesser understood subclass of performative commands that are the focus of this work. Utterances like ``The five of hearts is in the top corner" do not even encode an action with respect to the object mentioned, let alone an agent, but can nevertheless be used to elicit action of an addressee in certain contexts. 

As a motivating example, consider the following exchange between two players describing their respective hands:

\begin{displayquote}
P1: 3h, 4h and ks \\
P2: i have a queen of diamonds and ace of club \\
P2: we have a  mess lol
\end{displayquote}

\noindent Despite Player 2's concerns, a strategy emerges shortly thereafter when Player 1 finds an additional hearts card:

\begin{displayquote}
P1: i have 3h,4h,6h \\
P2: ok so we need to collect hearts then 
\end{displayquote}

\noindent At this point in the transcript, all that has been committed to the common ground is that Player 1 has a full hand of three proximal hearts cards that could be relevant to a winning strategy, and Player 2 has two non-hearts cards. This is the very next utterance in the exchange:

\begin{displayquote}
P1: there is a 5h in the very top left corner
\end{displayquote}

\noindent Player 2 is seen immediately hereafter to navigate to the top left corner, pick up the five of hearts, and confirm:

\begin{displayquote}
P2: ok i got it :)
\end{displayquote}

\noindent In this exchange, Player 2 appears to understand not only that the five of hearts is relevant to the winning strategy of six consecutive hearts, but also that it makes more sense for her to act on information about its location than it does for Player 1 to do so.

\begin{figure*}
		\begin{subfigure}[t]{0.25\textwidth}
			        \centering
	         \includegraphics[height=1.3in]{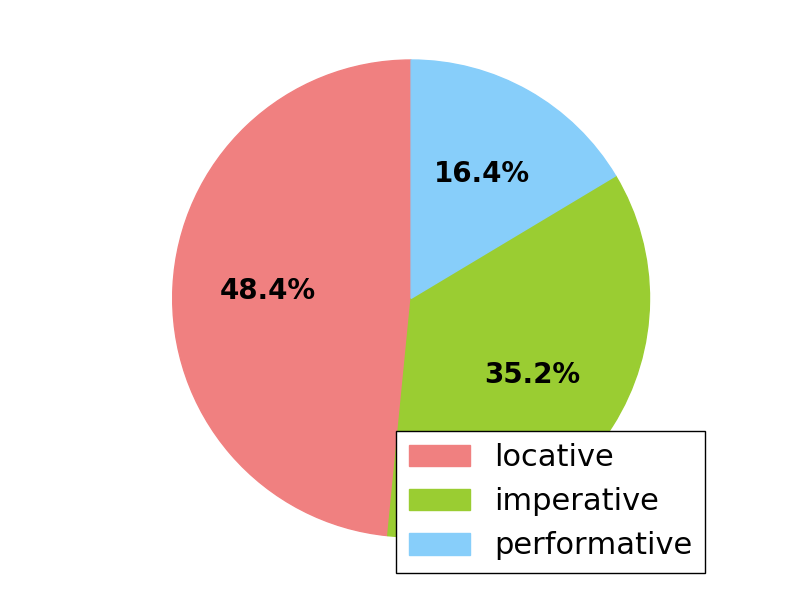}
	         \caption{pickup}
        \end{subfigure}%
		~ 
		\begin{subfigure}[t]{0.25\textwidth}
			\centering
			\includegraphics[height=1.3in]{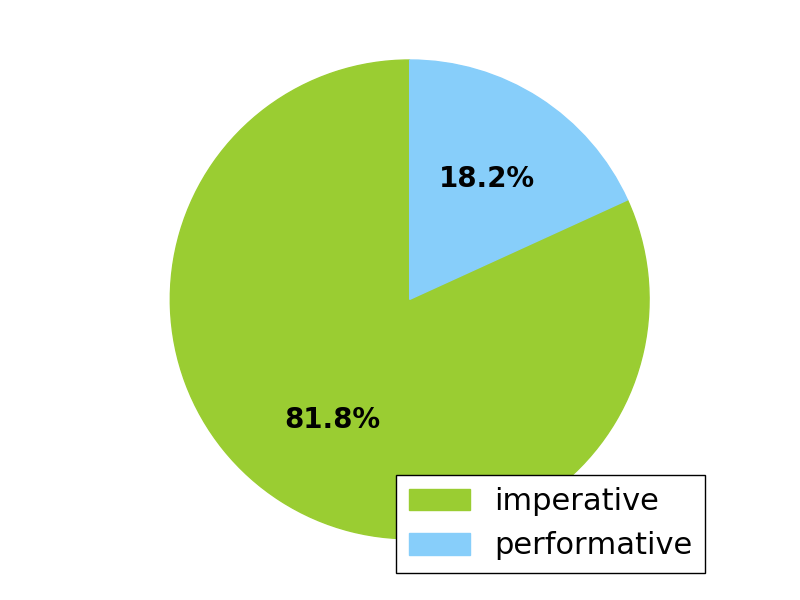}
			\caption{drop}
		\end{subfigure}%
		~
		\begin{subfigure}[t]{0.25\textwidth}
				\centering
					\includegraphics[height=1.3in]{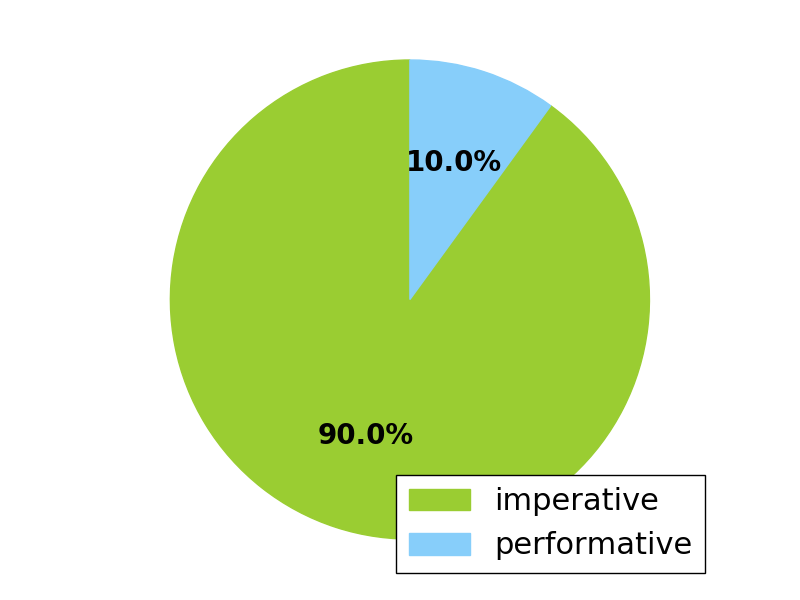}
				\caption{conversation}
		\end{subfigure}%
			\begin{subfigure}[t]{0.25\textwidth}
		\centering
		\includegraphics[height=1.3in]{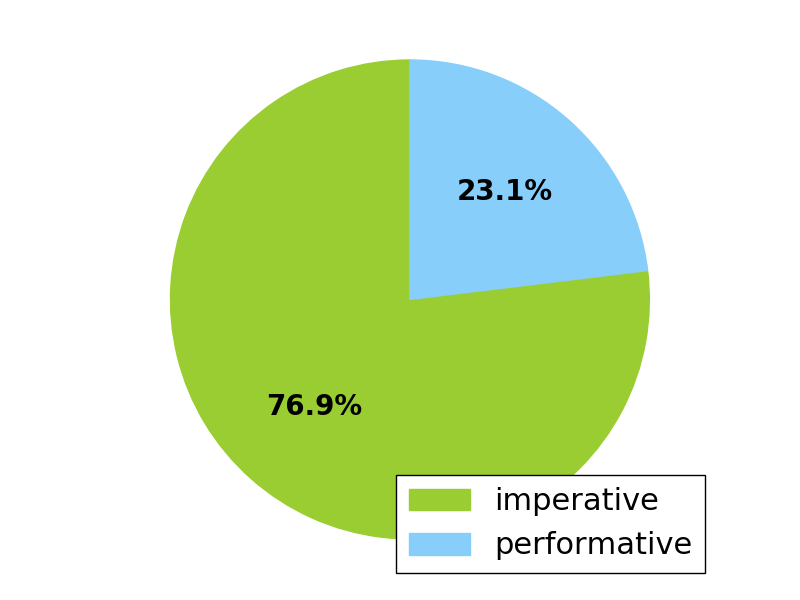}
		\caption{search}
	\end{subfigure}%
\caption{For each action, the distribution of command types}
\end{figure*}

This discourse encapsulates the collaborative reasoning pattern described by \cite{perrault1980plan}. The speaker assumes that the addressee is a cooperative agent. Thus, the addressee shares in the goal of attaining a winning game state and will act in a way so as to realize that goal. Recognizing the fact that the speaker would have to drop cards relevant to the goal to pick up the card at issue, a cooperative addressee will infer that she \emph{should} act by picking it up instead. In this way, locative utterances can be indirectly used as commands.

The distribution of command types for a subset of actions (\emph{pickup}, \emph{drop}, \emph{conversation}, and \emph{search}) is displayed in Figure 1. As depicted, for the majority of actions compelled by a speaker and taken by an addressee, the imperative is the predominant command strategy, followed by non-locative performatives. However locative commands appear to be the dominant strategy for eliciting card pickups in the corpus, constituting nearly half of all such commands observed.

This pattern demonstrates that for certain kinds of actions, it is quite natural to use the least direct, most context-dependent command strategy to elicit action of an addressee.

\section{Common Ground Effects on the Directive Force of Locative Utterances}
    We seek to understand how the discourse context of a game can influence the interpretation of locative utterances as commands. We therefore construct a binary classification task whereby we test how the role of a locative utterance can be resolved in context, evidenced by the actions that are taken as follow-ups to the utterance. In our task, one label denotes addressee follow-up in the form of acting to pick up the card in question, signaling her intention to act, or asking a clarifying question about its whereabouts. The second label denotes that either the speaker acts on their own utterance or neither agent does.  

\subsection{Annotation Details}
    Using a random sample of 200 transcripts from the corpus, we identify instances where a locative utterance is made and we annotate the common ground up to this utterance. This yields 55 distinct transcripts constituting 94 utterances with this particular phenomenon. 
    
    Our common ground annotations include the following information about the game state as indicated by players' utterances: cards in the players hands, player location, known information about the existence or location of cards, strategic statements made by players about needed cards, and whether a player is able to act with respect to an at-issue card.

\subsection{Experiments}
     Our aim in devising this task is to investigate connection between common ground knowledge and the illocutionary effects of locative utterances. We therefore train a standard logistic regression classifier and experiment with a few carefully designed features that encode constraints on player action, and which should hypothetically trigger the interpretation of locative utterances as indirect commands. We experiment with the following features:
     
     \begin{itemize}  
        \item \textbf{Edit Distance}: We use the minimal number of edits for an optimal solution as a feature. Given the cards in the players' hands at a given point in the game, we can define an optimal solution based on the number of edits that must be made to the hands to achieve that optimal solution. An edit is defined as either picking up or dropping a card, and each such action has a cost of 1. An optimal solution is defined as the one that requires the minimal number of edits given the current hands. For example, if player 1 has a 2H, 3H, and 4H and player 2 has a 6H and a 7H, the optimal solution is the 2H, 3H, 4H, 5H, 6H, and 7H. Such a solution requires a single edit because player 2 simply has to pick up a 5H. This feature seeks to capture the intuition that an addressee should tend to act with respect to a card when the edit distance is not particularly high and hence the game is near a winning state.
        
        \item \textbf{Explicit Goal}: This binary feature is triggered in two cases: 1) When the suit of card mentioned matches the agreed-upon suit strategy in the common ground and 2) When the card mentioned appears in the set of cards the addressee claims to need. This models the prediction that locative utterances are more likely to be indirect commands when they are relevant to a well-defined goal.

         \item \textbf{Full Hands}: This binary feature is triggered when the speaker has three cards of the same suit as the card mentioned, and which are associated with some winning six-card straight, but the addressee does not. This models the prediction that locative utterances are likely to be indirect commands when they provide information relevant to winning, but only the addressee can act as such.
         
     \end{itemize}

Single-feature classifiers are compared against a number of baselines to help benchmark our predictive task. Our first baseline, which is context-agnostic, seeks to capture the intuition that the role of a locative utterance is entirely ambiguous when considered in isolation. This baseline predicts the agent follow-up using a Bernoulli distribution weighted according to the class priors of the training data.
    
The second baseline incorporates surface-level dialogue context via bigram features of all the utterance exchanged between players up to and including the locative utterance. We also experimented with a unigram baseline but found that its performance was inferior to that of the bigram.

\subsection{Results}
We test our common-ground features one at a time with our logistic regression model, as we are interested in seeing how successfully they encode agents' pragmatic inferences. We also combine the two best-performing common-ground features. 
We report the results of our experiments using an $F_1$ measure and a 0.8/0.2 train/test split of our data in Table 1.

 \begin{table}[t]
\small
\centering
\begin{tabular}{l|c}
Model & $F_1$ \\
\hline
\emph{Random} & 23.5 \\
\emph{Bigram} & 58.9 \\
Edit Distance & 62.5 \\
Explicit Goal & 76.2 \\
Full Hand & \textbf{82.3} \\
Explicit Goal + Full Hand & 77.7 \\
\end{tabular}
\caption{\label{results} $F_1$ performance as reported on the test set. Note our baselines are italicized.}
\end{table}

We see that of our two baselines, the bigram model performs better. This bigram model also uses 2,916 distinct lexical features which makes it a highly overspecified model for our moderate data size. 

We find that our single-feature context-sensitive models both significantly outperform our baselines. Our Explicit Goal feature outperforms the Edit Distance feature by about 14\%, which indicates that locative utterances are often interpreted as commands in the presence of an explicit, common goal. The Full Hands feature outperforms the Explicit Goal feature by about 6\%. This strongly suggests that constraints on speaker action play a role in determining the illocutionary effect of a locative utterance. An addressee of such an utterance will tend to act accordingly when their partner cannot pick up the card mentioned, and when the card in question brings them closer to winning the game. We find that combining the Explicit Goal and Full Hands features improved performance over only using the Explicit Goal feature but reduced overall performance. This could be because the two features encode some common information about the agents' pragmatic implicatures during the game, and hence their correlative effects tend to degrade the combined model performance.

\section{Conclusion}

In this work, we have performed an extensive study of command types as present in the Cards corpus. Using the corpus as a test bed for grounded natural language interaction among agents with a shared goal, we describe a variety of utterances that may function as indirect commands when regarded in context.  In particular, locative utterances, which are not conventionally associated with command interpretations, are shown to operate as commands when considered in relation to situational constraints in the course of collaborative interaction. We develop a predictive task to show that models with carefully-designed features incorporating game state information can help agents effectively perform such pragmatic inferences. 

\section{Acknowledgments}
The authors would like to thank Christopher Potts and all of the anonymous reviewers for their valuable insights and feedback.

\bibliographystyle{chicago}
\bibliography{commands}

\begin{thebibliography}{}

\bibitem[\protect\citeauthoryear{Allen and Perrault}{Allen and
  Perrault}{1980}]{allen1980analyzing}
Allen, J.~F. and C.~R. Perrault (1980).
\newblock Analyzing intention in utterances.
\newblock {\em Artificial intelligence\/}~{\em 15\/}(3), 143--178.

\bibitem[\protect\citeauthoryear{Austin}{Austin}{1975}]{austin1975things}
Austin, J.~L. (1975).
\newblock {\em How to do things with words}.
\newblock Oxford University Press.

\bibitem[\protect\citeauthoryear{Beun}{Beun}{2000}]{beun2000context}
Beun, R.-J. (2000).
\newblock Context and form: Declarative or interrogative, that is the question.
\newblock {\em Abduction, Belief, and Context in Dialogue: Studies in
  Computational Pragmatics\/}~{\em 1}, 311--325.

\bibitem[\protect\citeauthoryear{Bolt}{Bolt}{1980}]{bolt1980put}
Bolt, R.~A. (1980).
\newblock {\em “Put-that-there”: Voice and gesture at the graphics
  interface}, Volume~14.
\newblock ACM.

\bibitem[\protect\citeauthoryear{Chai, She, Fang, Ottarson, Littley, Liu, and
  Hanson}{Chai et~al.}{2014}]{chai2014collaborative}
Chai, J.~Y., L.~She, R.~Fang, S.~Ottarson, C.~Littley, C.~Liu, and K.~Hanson
  (2014).
\newblock Collaborative effort towards common ground in situated human-robot
  dialogue.
\newblock In {\em Proceedings of the 2014 ACM/IEEE international conference on
  Human-robot interaction}, pp.\  33--40. ACM.

\bibitem[\protect\citeauthoryear{Clark}{Clark}{1979}]{clark1979responding}
Clark, H.~H. (1979).
\newblock Responding to indirect speech acts.
\newblock {\em Cognitive psychology\/}~{\em 11\/}(4), 430--477.

\bibitem[\protect\citeauthoryear{Clark}{Clark}{1996}]{clarkusinglanguage1996}
Clark, H.~H. (1996).
\newblock {\em Using language}.
\newblock Cambridge: Cambridge University Press.

\bibitem[\protect\citeauthoryear{Condoravdi and Lauer}{Condoravdi and
  Lauer}{2012}]{condoravdi2012imperatives}
Condoravdi, C. and S.~Lauer (2012).
\newblock Imperatives: Meaning and illocutionary force.
\newblock {\em Empirical issues in syntax and semantics\/}~{\em 9}, 37--58.

\bibitem[\protect\citeauthoryear{Degen, Franke, and J{\"a}ger}{Degen
  et~al.}{2013}]{degen2013cost}
Degen, J., M.~Franke, and G.~J{\"a}ger (2013).
\newblock Cost-based pragmatic inference about referential expressions.
\newblock In {\em CogSci}.

\bibitem[\protect\citeauthoryear{Djalali, Clausen, Lauer, Schultz, and
  Potts}{Djalali et~al.}{2011}]{Djalali-etal:2011}
Djalali, A., D.~Clausen, S.~Lauer, K.~Schultz, and C.~Potts (2011, November).
\newblock Modeling expert effects and common ground using {Q}uestions {U}nder
  {D}iscussion.
\newblock In {\em Proceedings of the {AAAI} Workshop on Building
  Representations of Common Ground with Intelligent Agents}, Washington, DC.
  Association for the Advancement of Artificial Intelligence.

\bibitem[\protect\citeauthoryear{Djalali, Lauer, and Potts}{Djalali
  et~al.}{2012}]{Djalali:Lauer:Potts:2012}
Djalali, A., S.~Lauer, and C.~Potts (2012).
\newblock Corpus evidence for preference-driven interpretation.
\newblock In M.~Aloni, V.~Kimmelman, F.~Roelofsen, G.~W. Sassoon, K.~Schulz,
  and M.~Westera (Eds.), {\em Proceedings of the 18th {A}msterdam Colloquium:
  Revised Selected Papers}, Berlin, pp.\  150--159. Springer.

\bibitem[\protect\citeauthoryear{Kaufmann}{Kaufmann}{2016}]{kaufmann2016fine}
Kaufmann, M. (2016).
\newblock Fine-tuning natural language imperatives.
\newblock {\em Journal of Logic and Computation\/}, exw009.

\bibitem[\protect\citeauthoryear{Kaufmann and Schwager}{Kaufmann and
  Schwager}{2009}]{kaufmann2009unified}
Kaufmann, S. and M.~Schwager (2009).
\newblock A unified analysis of conditional imperatives.
\newblock In {\em Semantics and Linguistic Theory}, Volume~19, pp.\  239--256.

\bibitem[\protect\citeauthoryear{Malamud and Stephenson}{Malamud and
  Stephenson}{2015}]{malamud2015three}
Malamud, S.~A. and T.~Stephenson (2015).
\newblock Three ways to avoid commitments: Declarative force modifiers in the
  conversational scoreboard.
\newblock {\em Journal of Semantics\/}~{\em 32\/}(2), 275--311.

\bibitem[\protect\citeauthoryear{Perrault and Allen}{Perrault and
  Allen}{1980}]{perrault1980plan}
Perrault, C.~R. and J.~F. Allen (1980).
\newblock A plan-based analysis of indirect speech acts.
\newblock {\em Computational Linguistics\/}~{\em 6\/}(3-4), 167--182.

\bibitem[\protect\citeauthoryear{Portner}{Portner}{2007}]{portner2007imperatives}
Portner, P. (2007).
\newblock Imperatives and modals.
\newblock {\em Natural Language Semantics\/}~{\em 15\/}(4), 351--383.

\bibitem[\protect\citeauthoryear{Potts}{Potts}{2012}]{Potts:2012WCCFL}
Potts, C. (2012).
\newblock Goal-driven answers in the {C}ards dialogue corpus.
\newblock In N.~Arnett and R.~Bennett (Eds.), {\em Proceedings of the 30th
  {W}est {C}oast {C}onference on {F}ormal {L}inguistics}, Somerville, MA, pp.\
  1--20. Cascadilla Press.

\bibitem[\protect\citeauthoryear{Searle}{Searle}{1989}]{searle1989performatives}
Searle, J.~R. (1989).
\newblock How performatives work.
\newblock {\em Linguistics and philosophy\/}~{\em 12\/}(5), 535--558.

\bibitem[\protect\citeauthoryear{Tellex, Thaker, Deits, Kollar, and Roy}{Tellex
  et~al.}{2012}]{tellex2012toward}
Tellex, S., P.~Thaker, R.~Deits, T.~Kollar, and N.~Roy (2012).
\newblock Toward information theoretic human-robot dialog.
\newblock In {\em Robotics: Science and Systems}, Volume~2, pp.\ ~3.

\bibitem[\protect\citeauthoryear{Tellex, Kollar, Dickerson, Walter, Banerjee,
  Teller, and Roy}{Tellex et~al.}{2011}]{tellex2011understanding}
Tellex, S.~A., T.~F. Kollar, S.~R. Dickerson, M.~R. Walter, A.~Banerjee,
  S.~Teller, and N.~Roy (2011).
\newblock Understanding natural language commands for robotic navigation and
  mobile manipulation.

\bibitem[\protect\citeauthoryear{Vogel, G{\'o}mez~Emilsson, Frank, Jurafsky,
  and Potts}{Vogel et~al.}{2014}]{Vogel-etal:2014}
Vogel, A., A.~G{\'o}mez~Emilsson, M.~C. Frank, D.~Jurafsky, and C.~Potts (2014,
  July).
\newblock Learning to reason pragmatically with cognitive limitations.
\newblock In {\em Proceedings of the 36th Annual Meeting of the {C}ognitive
  {S}cience {S}ociety}, Wheat Ridge, CO, pp.\  3055--3060. Cognitive Science
  Society.

\bibitem[\protect\citeauthoryear{Vogel, Potts, and Jurafsky}{Vogel
  et~al.}{2013}]{Vogel:Potts:Jurafsky:2013}
Vogel, A., C.~Potts, and D.~Jurafsky (2013, August).
\newblock Implicatures and nested beliefs in approximate
  {D}ecentralized-{POMDPs}.
\newblock In {\em Proceedings of the 2013 Annual Conference of the
  {A}ssociation for {C}omputational {L}inguistics}, Stroudsburg, PA, pp.\
  74--80. Association for Computational Linguistics.

\bibitem[\protect\citeauthoryear{Walter, Antone, Chuangsuwanich, Correa, Davis,
  Fletcher, Frazzoli, Friedman, Glass, and How}{Walter
  et~al.}{2015}]{walter2015situationally}
Walter, M.~R., M.~Antone, E.~Chuangsuwanich, A.~Correa, R.~Davis, L.~Fletcher,
  E.~Frazzoli, Y.~Friedman, J.~Glass, and J.~P. How (2015).
\newblock A situationally aware voice-commandable robotic forklift working
  alongside people in unstructured outdoor environments.
\newblock {\em Journal of Field Robotics\/}~{\em 32\/}(4), 590--628.

\bibitem[\protect\citeauthoryear{Wierzbicka}{Wierzbicka}{1991}]{wierzbickacrossculturalpragmatics}
Wierzbicka, A. (1991).
\newblock {\em Cross-cultural pragmatics: the semantics of human interaction}.
\newblock Berlin: Mouton de Gruyter.

\end{thebibliography}

\end{document}